\documentclass[10pt]{article} 
\usepackage[preprint]{tmlr}

\usepackage{amsmath,amsfonts,bm}

\def\eqref#1{equation~\ref{#1}}

\def\1{\bm{1}}

\def\vmu{{\bm{\mu}}}

\def\vc{{\bm{c}}}

\def\vk{{\bm{k}}}

\def\vs{{\bm{s}}}

\def\vu{{\bm{u}}}

\def\vx{{\bm{x}}}
\def\vy{{\bm{y}}}

\DeclareMathAlphabet{\mathsfit}{\encodingdefault}{\sfdefault}{m}{sl}
\SetMathAlphabet{\mathsfit}{bold}{\encodingdefault}{\sfdefault}{bx}{n}

\usepackage{hyperref}
\usepackage{url}
\usepackage{graphicx}
\usepackage[noabbrev]{cleveref}
\usepackage{enumitem}
\usepackage{amsmath, amsthm, amssymb}
\usepackage[noend]{algorithmic}
\usepackage{booktabs}
\usepackage{algorithm}
\usepackage{url}
\usepackage{thmtools}
\usepackage{hyperref}
\usepackage{subcaption}

\newtheorem{remark}{Remark}[section]
\theoremstyle{definition}

\title{Correcting Boundary Bias and Observation Independence in Bayesian Experimental Design}

\author{\name Sanna Jarl \email sanna.jarl@angstrom.uu.se \\
      \addr Department of Material Science \& Engineering \\
      Uppsala University \\ RISE
      \AND
      \name Jens Sjölund \email jens.sjolund@it.uu.se \\
      \addr Department of Information Technology \\
      Uppsala University
      \AND
      \name Jonathan J. S. Scragg \email jonathan.scragg@angstrom.uu.se \\
      \addr Department of Material Science \& Engineering \\
      Uppsala University
      \AND
      \name Maria Bånkestad \email maria.bankestad@ri.se \\
      \addr RISE
      }

\def\month{MM}  
\def\year{YYYY} 
\def\openreview{\url{https://openreview.net/forum?id=XXXX}} 

\begin{document}

\maketitle

\begin{abstract}
In many experimental settings, active learning can improve sample efficiency by sequentially selecting where to measure, which is particularly valuable when experiments are expensive.
Gaussian processes with variance-based acquisition criteria are widely used for this purpose, but have two limitations. 
First, they are observation-independent: their posterior variance depends only on \emph{where} samples are acquired, not on \emph{what} is measured, impairing their sensitivity to the structure of the acquired data.  
Second, they inflate the variance near boundaries, leading to excessive sampling at the edges of the space compared to the interior. These limitations undermine the  gains in sampling efficiency expected from sequential acquisition. 
We address both limitations. 
We derive a reconstruction-driven design density and use the posterior mean to build a training-free warp that places more measurements where the target function varies rapidly. 
A geometric equalizer separately corrects boundary bias.
Across sixteen synthetic and two real-data benchmarks, the geometric equalizer consistently improves function reconstruction by correcting boundary bias, while the reconstruction warp provides further gains by concentrating measurements where the posterior mean varies rapidly.
\end{abstract}

\section{Introduction}
\label{sec:introduction}

Experiments in the physical sciences are often expensive.
When only a limited number can be afforded, it is natural to choose them \textit{sequentially}, with the outcomes of earlier experiments guiding the next choice.
In many settings, as opposed to seeking an optimum, the goal is to reconstruct how an outcome varies across the design space~\citep{abolhasani2023rise}. This is an active-learning task~\citep{settles2009active,di2024active} rather than a Bayesian-optimization problem~\citep{garnett2023bayesian}.

A common approach is to model the unknown function with a Gaussian process (GP)~\citep{rasmussen2006gaussian}, score candidate experiments by predictive uncertainty, and select the one with the largest score.
For variance-based GP criteria, this score depends on the observation locations and kernel hyperparameters, but not directly on the observed values~\citep{mackay1992information,krause2008near}: it is \emph{observation-independent}.
Two regions with the same observation spacing therefore receive an identical score, even if the function varies smoothly in one and sharply in the other~\citep{belakaria2024active}.
This limitation matters especially for nonstationary functions, where complexity may be concentrated near transitions and sharp features~\citep{gramacy2009adaptive,crombecq2011novel}.
In such cases, geometric coverage alone cannot allocate additional measurements to regions that require finer resolution.
With a stationary GP, observed values can affect the variance only through hyperparameter refitting, but this adaptation is global. Often, a single lengthscale cannot account for the fact that some regions require much finer resolution than others~\citep{heinonen2016non}.\looseness-1

A second problem appears on bounded domains, where points near the boundary have a truncated correlation neighborhood that makes their posterior variance systematically higher.
This creates \emph{boundary bias}: variance-based acquisition favors the boundary independently of the observed values, with the effect becoming more pronounced as dimension increases~\citep{bankestad2026boundary}.
This boundary bias can be reduced by modifying the kernel, using integrated criteria, or directly correcting the acquisition score.
While kernel corrections can calibrate variance near the boundary~\citep{bankestad2026boundary}, they do not change the fact that a boundary observation informs less of the domain.
Integrated criteria such as negative integrated posterior variance (NIPV) and expected predictive information gain (EPIG)~\citep{smith2023prediction} account for this geometry but require a domain integral at every round.
We instead correct this boundary bias using the expected posterior variance under reference space-filling designs.
The resulting \emph{geometric equalizer} depends only on the domain geometry, kernel, noise level, and design size.
\looseness-1

\begin{figure}
\centering
\includegraphics[width=\linewidth]{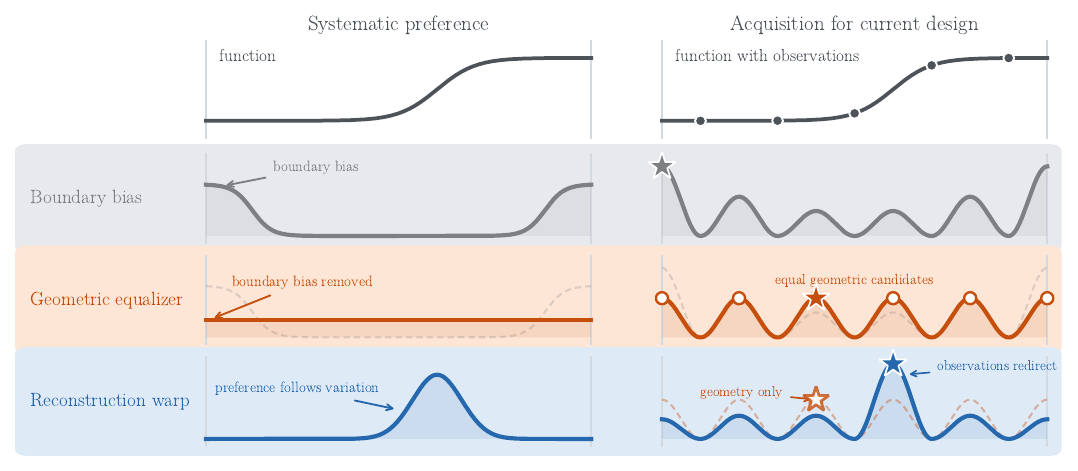}
\caption{Schematic overview of the acquisition geometry. The top row shows the latent function that we aim to reconstruct, with and without five evenly distributed observations. The rows below show the acquisition functions in the standard variance-based case (gray), exhibiting boundary bias; after correcting this bias with the geometric equalizer (orange); and, finally, after the reconstruction warp additionally redirects acquisition to regions where the posterior mean varies rapidly (blue). Stars indicate the selected next points.}
\label{fig:warp_overview}
\end{figure}

The geometric equalizer does not use the observations. To make acquisition observation-dependent, we derive a reconstruction-driven design density that allocates points where the function varies rapidly. For an unknown function, we use the posterior mean to construct a training-free \emph{reconstruction warp} that redirects acquisition to such regions. \Cref{fig:warp_overview} illustrates both limitations and their effects on sequential experiment design.
Importantly, both corrections act only on acquisition, while the prediction GP remains unchanged.

To evaluate observation independence and boundary bias separately, we first turn to a boundary-free sphere.
There, covariance-based acquisition closely tracks farthest-point sampling (FPS), and we use the reconstruction warp to test whether resolved function variation can improve measurement placement.
We then restore the bounded domain, where boundary bias can make variance maximization worse than an experiment design fixed in advance.
After correcting the boundary bias, we test the effect of observation-dependent warping.

In summary, our contributions are:
\begin{enumerate}[label=(\roman*), leftmargin=2em, itemsep=2pt, topsep=2pt]
\item We show that on a boundary-free domain, EIG, NIPV, and EPIG closely track farthest-point sampling. Under fixed hyperparameters, these acquisition rules are exactly open-loop, and per-round hyperparameter refitting changes little (\Cref{sec:baselines_reveal}).
\item We show that, at small budgets, boundary bias can make sequential variance maximization worse than a design fixed in advance (\Cref{sec:cube_flip}).
\item We derive a \emph{geometric equalizer} that corrects boundary bias using the expected posterior variance under reference space-filling designs (\Cref{sec:boundary_calibration,sec:cube}).
\item We derive a reconstruction-driven design density and use the posterior mean to construct an observation-dependent, training-free \emph{reconstruction warp} that directs measurements toward regions where the function varies rapidly (\Cref{sec:design_density,sec:cube}).
\end{enumerate}

\section{Preliminaries}
\label{sec:preliminaries}

Let $z \sim \mathcal{GP}(m(\vx), k_\theta(\vx, \vx'))$ denote a Gaussian process prior with mean function $m(\vx)$ and covariance kernel $k_\theta(\vx, \vx')$.
Given observations $\mathcal{D} = \{X, \vy\}$ with $y_i = z(\vx_i) + \varepsilon_i$,
$\varepsilon_i \sim \mathcal{N}(0, \sigma_n^2)$, the predictive posterior at $\vx_*$ is
$p(z_* \mid \mathcal{D}, \vx_*) = \mathcal{N}(\mu(\vx_*), s^2(\vx_*))$
with
\begin{align}
\mu(\vx_*) &= m(\vx_*) + \vk_*^\top (K + \sigma_n^2 I)^{-1}(\vy - m(X)), \label{eq:gp_mean}\\
s^2(\vx_*) &= k_\theta(\vx_*, \vx_*) - \vk_*^\top (K + \sigma_n^2 I)^{-1} \vk_*, \label{eq:gp_var}
\end{align}
where $K_{ij} = k_\theta(\vx_i, \vx_j)$ and $\vk_* = (k_\theta(\vx_1, \vx_*), \ldots, k_\theta(\vx_N, \vx_*))^\top$.

\begin{remark}[Observation independence of variance-based acquisition]
    \label{prop:y_independence}
    Under a GP with fixed hyperparameters $\theta$, the posterior variance $s^2(\vx_*)$ and, hence, the expected information gain
    \begin{equation}
        \mathrm{EIG}(\vx_*) = \frac{1}{2}\log \left(\frac{s^2(\vx_*)+\sigma_n^2}{\sigma_n^2}\right)
    \label{eq:eig_gp}
    \end{equation}
    are conditionally independent of the observed values $\vy$ given the input locations $X$.
\end{remark}
This follows immediately from \eqref{eq:gp_var}: $s^2(\vx_*)$ depends on $X$ and $\theta$ but not on $\vy$. (See Appendix~\ref{app:eig_derivation} for the derivation of $\text{EIG}(\vx_*)$.)
Although this observation is well known~\citep[e.g.,][]{krause2008near}, its consequence for sequential experimental design is easy to overlook.
Since the integrated criteria NIPV~\citep{seo2000nipv} and EPIG~\citep{smith2023prediction} are functions of the covariance, they are also observation-independent (Appendix~\ref{app:zoo_criteria}).\looseness-1

With fixed hyperparameters and deterministic tie-breaking, observation-independent acquisition rules are \emph{open-loop}: they generate the same sequence of experiments whether run sequentially, observing each $y$ before the next selection, or computed before any function value is revealed.
The reconstruction warp closes the feedback loop by introducing a dependence on the posterior mean.

\section{Related Work}
\label{sec:related_work}
Gaussian-process experimental design commonly selects observations by posterior uncertainty~\citep{mackay1992information,krause2008near}. Pointwise variance and EIG favor uncertain locations, while integrated criteria such as NIPV~\citep{seo2000nipv} and EPIG~\citep{smith2023prediction} weight a candidate by its influence on predictions across the domain.  Classical links between maximum-entropy GP designs and maximin distance designs~\citep{johnson1990minimax,sacks1989design} connect this covariance-based view to geometric coverage.

One route to observation dependence is to enrich the surrogate. Learned input warps~\citep{kumar_warp,balandat2020botorch}, manifold GPs~\citep{calandra2016manifold}, nonstationary kernels~\citep{heinonen2016non}, and deep GP surrogates~\citep{sauer2023active} let observations reshape the covariance used for both prediction and acquisition (hyperparameter fitting being a special case). Because these approaches modify prediction and acquisition together, it is difficult to separate their effects. We instead decouple prediction and acquisition, applying our corrections only to the latter. Related ideas appear in surrogate modulation~\citep{bodin2020modulating}, local-complexity methods such as LOLA--Voronoi~\citep{crombecq2011novel}, and quantization, where sampling density follows local approximation error~\citep{deboor,gersho1979asymptotically}. Our reconstruction warp instead starts from a reconstruction-driven design density and uses it to reshape the acquisition geometry.\looseness-1

For broader reviews of Bayesian experimental design and sequential surrogate modeling, see \citet{rainforth2024modern,gramacy2020surrogates}; per-method details are in Appendix~\ref{app:extended_related_work}.

\section{Method}
\label{sec:method}

The two problems identified above require separate measures. The geometric equalizer corrects boundary bias, while the reconstruction warp introduces observation dependence by using the posterior mean to allocate more measurements where the function varies rapidly.

\subsection{Decoupling prediction and acquisition}
\label{sec:decoupled_framework}

We intentionally use separate GPs for prediction and acquisition, conditioned on the same observations.
The \emph{prediction GP} has hyperparameters $\theta_{\mathrm p}$ fitted as usual and is not affected by either of our corrections.
The \emph{acquisition GP} is used only to select the next experiment, with hyperparameters $\theta_{\mathrm a}$ kept fixed to isolate the effect of the acquisition rule.
Both corrections act only on acquisition; the prediction GP remains unchanged.
Keeping prediction and acquisition separate also lets us evaluate the selected experiments independently of the model that selected them (\Cref{sec:experiments}).

We use EIG for acquisition. With fixed observation noise, EIG is a strictly increasing function of the posterior variance, so maximizing EIG is equivalent to maximizing $s^2(\vx)$. We therefore work directly with the variance.

\subsection{Geometric equalizer}
\label{sec:boundary_calibration}

The geometric equalizer corrects the boundary bias caused by truncated correlation neighborhoods.
To quantify this boundary bias, we average the posterior variance over reference space-filling designs.
Let $Q_N$ denote a reference distribution over designs of size $N$, implemented here with scrambled Sobol designs~\citep{Owen1998ScramblingSA}, and define
\begin{equation}
    \bar v_N(\vx)
    =
    \mathbb{E}_{X_N \sim Q_N}
    \big[s^2(\vx \mid X_N)\big],
    \label{eq:reference_variance}
\end{equation}
where $s^2(\vx \mid X_N)$ is the posterior variance of the acquisition GP conditioned on $X_N$.
The reference variance is independent of the observed values but depends on the domain, kernel, noise level, and design size.
On a bounded domain, $\bar v_N(\vx)$ is nearly flat in the interior and increases near the boundary, where the correlation neighborhood is truncated.

\begin{figure}[t]
    \centering
    \includegraphics[width=\linewidth]{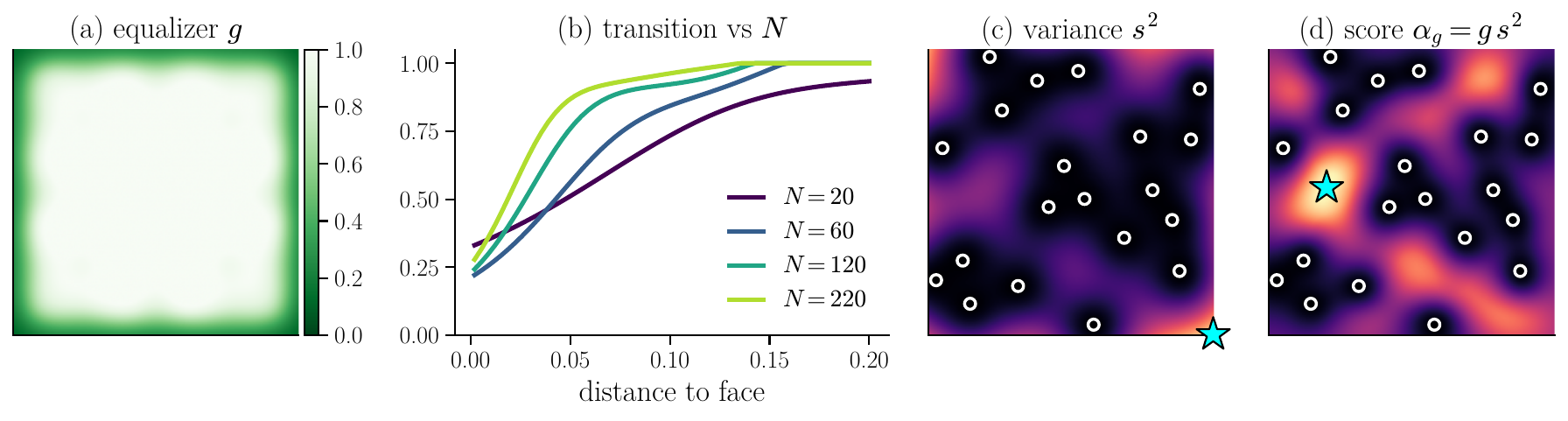}
    \caption{
    (a) The geometric equalizer in 2D, equal to one in the interior and decreasing near the boundary.
    (b) The transition depends on the design size $N$ and the acquisition lengthscale $\ell_{\mathrm a}$.
    Without equalization, (c), the raw variance maximum lies in a corner; after equalization, (d), the selected point moves into the interior.\looseness-1}
    \label{fig:map}
\end{figure}

We use $\bar v_N$ to downweight the acquisition score near the boundary.
We take the domain center as the interior reference,
$\bar v_{\mathrm{int}}=\bar v_N(\tfrac12,\ldots,\tfrac12)$, and define
\begin{equation}
    g(\vx;\ell_{\mathrm a},N)
    =
    \min\!\Big(
        1,\,
        \frac{\bar v_{\mathrm{int}}}{\bar v_N(\vx)}
    \Big),
    \qquad
    \alpha_g(\vx)
    =
    g(\vx;\ell_{\mathrm a},N)s^2(\vx).
    \label{eq:geo_equalizer}
\end{equation}
Because $\bar v_N$ is estimated numerically, small fluctuations in the nearly flat interior can make the ratio
$\bar v_{\mathrm{int}}/\bar v_N(\vx)$ vary around one even where there is little boundary effect to correct.
Clipping at one suppresses these fluctuations while retaining the correction where $\bar v_N$ rises near the boundary.
\Cref{fig:map} shows both the shape of the equalizer and its effect on acquisition: the correction is strongest near the boundary, where it can move the variance maximum away from a corner and into the interior.

For a fixed domain, kernel family, and noise level, the equalizer depends only on position, lengthscale, and design size, not on the observed values, so it can be precomputed and evaluated cheaply during acquisition.
Reflection and permutation symmetries let us tabulate it once per dimension and evaluate it by table lookup (\Cref{fig:cube_ell}e).
Appendix~\ref{app:selection_bias} uses a function-free diagnostic to show that variance maximization places many selections near faces and corners, whereas the equalizer moves them toward the interior, closer to integrated criteria such as NIPV and EPIG.\looseness-1

Boundary bias can also be reduced by modifying the kernel.
The Neumann acquisition kernel of \citet{bankestad2026boundary} completes the truncated correlation neighborhood using reflected images, which calibrates the posterior variance near the boundary.
This changes the acquisition variance but not how much of the domain a boundary observation informs; \Cref{sec:cube_flip} compares the two approaches empirically.

The geometric equalizer addresses boundary bias but remains independent of the observed values.
The reconstruction warp adds observation dependence.

\subsection{The reconstruction warp: introducing observation dependence}
\label{sec:design_density}

We use the posterior mean to determine where finer sampling resolution is needed. We first derive a design density from a local approximation to nearest-neighbor reconstruction error, and then use it to construct the target density for the training-free \emph{reconstruction warp}.\looseness-1

Assume that reconstruction error is integrated uniformly over a \(D\)-dimensional domain \(\mathcal X\).
Consider a design of \(N\) observation locations and, for fixed \(N\), represent the design by a normalized density \(p(\vx)\),
\begin{equation}
    \int_{\mathcal X} p(\vx)\,d\vx = 1.
\end{equation}
When the design is sufficiently dense for this continuum approximation to be meaningful, the point intensity at \(\vx\) is \(N p(\vx)\).
The typical distance from \(\vx\) to its nearest design point therefore scales as
\begin{equation}
    d(\vx) \propto [N p(\vx)]^{-1/D}.
\end{equation}
We further assume that \(f\) is locally well approximated by its first-order expansion over the nearest-neighbor distance.
Under this approximation, the local squared reconstruction error to the 1-NN $\vx_{\mathrm{nn}}$ scales as
\begin{equation}
    \bigl(f(\vx_{\mathrm{nn}})-f(\vx)\bigr)^2
    \propto
    \|\nabla f(\vx)\|^2 d(\vx)^2
    \propto
    N^{-2/D}
    \|\nabla f(\vx)\|^2 p(\vx)^{-2/D}.
\end{equation}

Integrating this local error over the domain and dropping the factor \(N^{-2/D}\), which does not depend on \(p\), gives the spatial allocation problem 
\begin{equation} 
\min_{p} \int_{\mathcal X} \|\nabla f(\vx)\|^2 p(\vx)^{-2/D}\,d\vx, \qquad \text{subject to } \int_{\mathcal X} p(\vx)\,d\vx = 1. \label{eq:density_objective} 
\end{equation}
Introducing a Lagrange multiplier \(\lambda\), pointwise stationarity gives
\begin{equation}
    -\frac{2}{D}
    \|\nabla f(\vx)\|^2
    p(\vx)^{-(D+2)/D}
    +\lambda
    =0,
\end{equation}
and therefore
\begin{equation}
p^*(\vx)
\propto
\|\nabla f(\vx)\|^{\frac{2D}{D+2}}.
\label{eq:opt_density}
\end{equation}
Thus \(N\) sets the overall sampling resolution, while \(\|\nabla f(\vx)\|\) determines how measurements are distributed across the domain: regions where \(f\) varies more rapidly receive more points.
Related density laws arise in optimal knot placement and asymptotic quantization~\citep{deboor,gersho1979asymptotically}.

The true gradient is unavailable during acquisition, so we estimate its spatial pattern from the posterior mean \(\mu_{\mathrm a}\) of the acquisition GP.
A small posterior-mean gradient is ambiguous: the function may be genuinely flat, or the region may be poorly observed and the posterior mean may have reverted toward its prior mean.
In the latter case, assigning little density to the region could prevent the estimate from improving.

\begin{figure}[t]
    \centering
    \includegraphics[width=\linewidth]{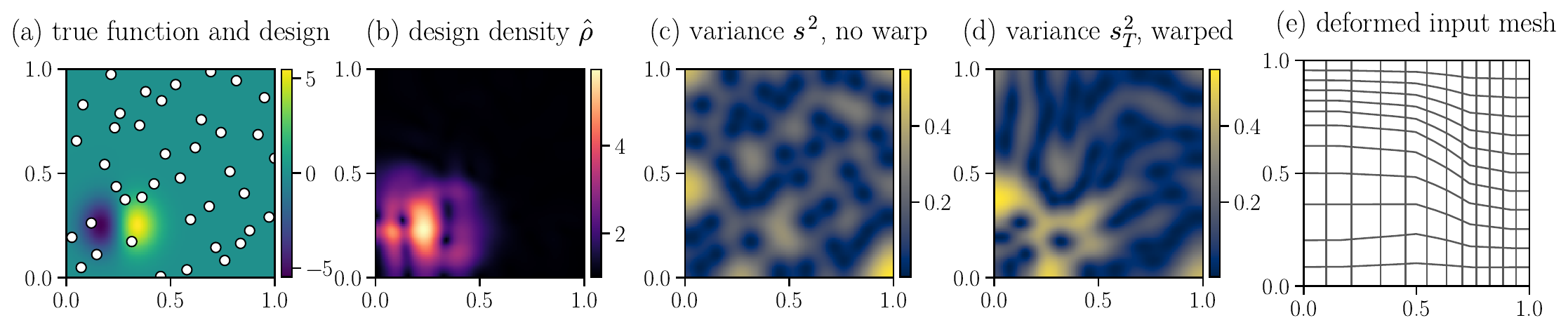}
    \caption{Reconstruction warp on the Gramacy--Lee benchmark ($D{=}2$, $N{=}40$).
    (a) The true function and current design.
    (b) The target density $\hat\rho$ in \eqref{eq:target_density}, constructed from the posterior-mean gradient and larger where the posterior mean varies rapidly.
    (c) Unwarped acquisition variance, driven mainly by geometric coverage and largest near the boundary.
    (d) Acquisition variance after applying the reconstruction warp, shifted toward regions of larger variation.
    (e) The corresponding deformation of the input coordinates.
    The reconstruction warp changes only the acquisition geometry; the prediction GP remains unchanged.\looseness=-1}
    \label{fig:warp_anatomy}
\end{figure}

We therefore retain a uniform component and define the unnormalized target density
\begin{equation}
    \hat\rho(\vx)
    =
    \rho_0
    +
    \left(
        \frac{\|\nabla\mu_{\mathrm a}(\vx)\|}{c}
    \right)^{\frac{2D}{D+2}},
    \label{eq:target_density}
\end{equation}
where \(c\) is a constant. 
The gradient-dependent term increases the target density where the posterior mean varies rapidly, while the uniform component limits this concentration and prevents poorly resolved regions from being neglected.
Normalizing by \(c\) makes the construction depend on the spatial pattern of the posterior-mean gradient rather than its overall scale.
If the posterior mean is exactly flat, the gradient-dependent term vanishes and \(\hat\rho\) is uniform.
\Cref{fig:warp_anatomy} illustrates the resulting chain from posterior-mean gradient to target density, coordinate deformation, and acquisition variance.

We normalize the target density to obtain
\begin{equation}
    \pi(\vx)
    =
    \frac{\hat\rho(\vx)}
    {\int_{\mathcal X}\hat\rho(\vu)\,d\vu}.
    \label{eq:normalized_target_density}
\end{equation}
The reconstruction warp is a transport map \(T\) satisfying
\begin{equation}
    T_\#\pi
    =
    \mathrm{Unif}([0,1]^D).
    \label{eq:transport}
\end{equation}
Equivalently, \(T^{-1}\) maps uniformly distributed points in the warped domain to points distributed according to \(\pi\) in the original domain.
Regions assigned higher target density are therefore expanded in the warped coordinates, so variance-based acquisition allocates finer resolution to them.
We construct \(T\) as a Knothe--Rosenblatt transport map from numerical estimates of the marginal and conditional CDFs.

The warp modifies only the acquisition kernel,
\begin{equation}
    k_{\theta_{\mathrm a},T}(\vx,\vx')
    =
    k_{\theta_{\mathrm a}}
    \bigl(T(\vx),T(\vx')\bigr).
    \label{eq:warped_kernel}
\end{equation}

Both candidate and observed locations are transformed when computing the acquisition variance, which we denote by \(s_T^2\).
The prediction GP remains unchanged.
The target density shapes the acquisition geometry rather than prescribing a fixed sampling pattern; the next point is still selected by maximizing the acquisition variance.
The map is rebuilt directly from the current posterior mean at each acquisition round and requires no inner optimization.
For an exactly flat posterior mean, \(\pi\) is uniform, \(T\) is the identity, and acquisition reduces to the unwarped case.

\subsection{Procedure}
\label{sec:procedure}

\Cref{alg:warped_active_learning} combines the geometric equalizer and reconstruction warp in a sequential acquisition loop.
\let\AND\undefined
\begin{algorithm}[ht]
\small
   \caption{Sequential acquisition with the geometric equalizer and reconstruction warp}
   \label{alg:warped_active_learning}
   {\bfseries Input:} Domain $\mathcal{X}$, noisy oracle $f$, prediction GP, acquisition GP with fixed hyperparameters $\theta_{\mathrm a}$, initial design size $b_0$, budget $B$.\\
   {\bfseries Output:} Dataset $\mathcal{D}_B$, fitted prediction GP.
\begin{algorithmic}[1]
   \STATE Initialize $\mathcal{D}_{b_0}$ with $b_0$ observations.
   \FOR{$i = b_0+1, \ldots, B$}
       \STATE Fit the prediction-GP hyperparameters $\theta_{\mathrm p}$ to $\mathcal{D}_{i-1}$ via MAP. \hfill \textit{// prediction}
       \STATE Condition the acquisition GP on $\mathcal{D}_{i-1}$ and evaluate $\nabla\mu_{\mathrm a}$ at $M$ Sobol probes.
       \STATE Form $\hat\rho$ using \eqref{eq:target_density} and construct the transport map $T$ such that $T_\#\pi=\mathrm{Unif}([0,1]^D)$. \hfill \textit{// reconstruction warp}
       \STATE Look up $g(\cdot\,;\ell_{\mathrm a},i-1)$ for the current design size. \hfill \textit{// geometric equalizer}
       \STATE Select
       $\displaystyle
       \vx_* \leftarrow
       \arg\max_{\vx\in\mathcal X}
       g(\vx;\ell_{\mathrm a},i-1)\,
       s_T^2(\vx\mid\mathcal D_{i-1}).
       $
       \STATE Observe $y_*=f(\vx_*)$ and set
       $\mathcal{D}_i=\mathcal{D}_{i-1}\cup\{(\vx_*,y_*)\}$.
   \ENDFOR
\end{algorithmic}
\end{algorithm}

We place a log-normal prior on the prediction-GP lengthscale~\citep{hvarfner2024vanilla}.
The additional per-round cost comes from evaluating the acquisition-GP posterior-mean gradient at the Sobol probes, constructing the numerical CDFs for the reconstruction warp, and looking up the geometric equalizer.\looseness-1

\section{Experiments}
\label{sec:experiments}

We evaluate the two effects and their corrections in stages. We first remove the boundary to isolate observation dependence, then restore the bounded domain to examine boundary bias. We then correct boundary bias with the geometric equalizer, and finally investigate the effect of the reconstruction warp once the geometric bias is removed.
The code to reproduce the experiments is available here\footnote{added upon acceptance}.

All acquired experiment designs are evaluated with the same $1$-nearest-neighbor reconstruction rule against a fixed reference target. The mean-square error of the nearest neighbor reconstruction on a dense, fixed, held out set is measured (details in Appendix \ref{app:evaluator}).
Performance is summarized by area reduction ($\mathrm{AR}$), the reduction in reconstruction error accumulated over acquisition rounds relative to FPS; positive values indicate improvement (Appendix~\ref{app:area_reduction}).

Unless stated otherwise, all GP-based acquisition methods use the same fixed Mat\'ern-$5/2$ acquisition kernel, with the lengthscale determined from design spacing and observation noise fixed at its known value.
We consider sixteen functions on bounded domains in two to six dimensions, and two real-data benchmarks (Appendix~\ref{app:test_functions}), each over $200$ acquisition rounds.
We used the three smooth box functions during method development; the remaining fifteen benchmarks were held out.
Results are averaged over ten seeds, with full configurations given in Appendix~\ref{app:configurations}.

Since the design-density derivation is based on nearest-neighbor reconstruction under squared error, the primary evaluator matches the loss used in that derivation.
As a robustness verification, Appendix~\ref{app:evaluator} reevaluates every acquired experiment design using $3$-NN, $5$-NN, and GP reconstruction; the main qualitative conclusions persist across reconstruction rules.

\subsection{A boundary-free domain: the sphere}
\label{sec:baselines_reveal}

On a bounded domain, observation independence and boundary bias act together.
Focusing on a boundary-free domain, namely the sphere, allows us to isolate the effect of observation-dependent acquisition.

\begin{figure}[t]
    \centering
    \includegraphics[width=\linewidth]{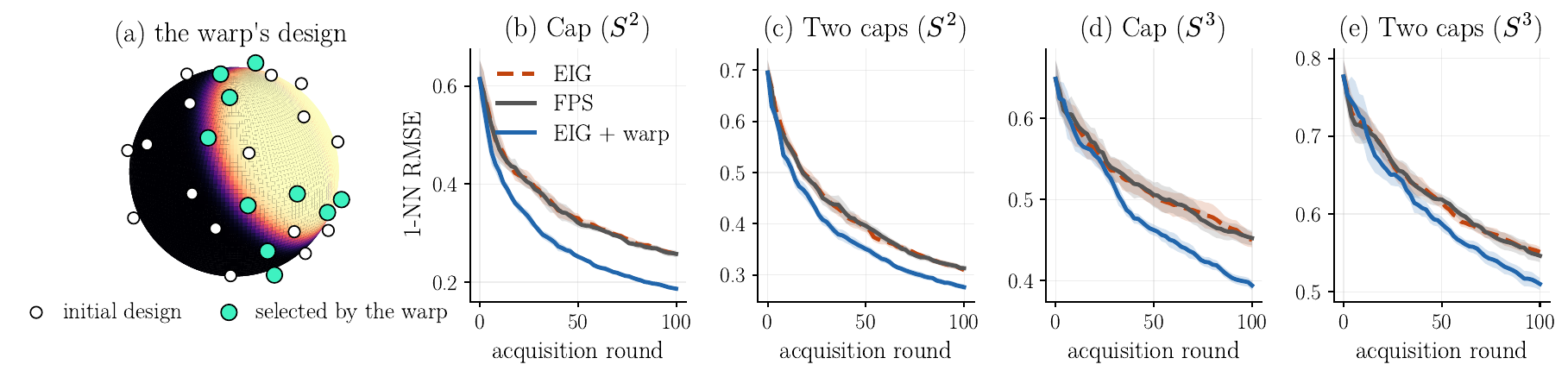}
    \caption{
    (a) The $S^2$ cap, with the initial design in white and points selected by the reconstruction warp in teal, concentrated around the transition ring.
    (b--e) At the same number of acquisition rounds, the reconstruction warp improves on FPS from early rounds on $S^2$, whereas on $S^3$ the advantage emerges later (mean and $95\%$ CI over $10$ seeds).}
    \label{fig:sphere_warp}
\end{figure}

The target functions are spherical caps on $S^2$ and $S^3$, spherical analogs of the smooth box, with flat regions separated by a narrow transition band (\Cref{fig:sphere_warp}(a)). Appendix~\ref{app:sphere} contains details of the experiments.
The GP uses a chordal Mat\'ern kernel, which remains positive definite on the sphere~\citep{gneiting2013strictly}.

\begin{table}[t]
\centering
\footnotesize
\setlength{\tabcolsep}{3pt}
\begin{tabular}{lcccccccc}
\hline
 & \multicolumn{4}{c}{$\mathrm{AR}$ (\%) $\uparrow$, 20 rounds} & \multicolumn{4}{c}{$\mathrm{AR}$ (\%) $\uparrow$, 100 rounds} \\
Benchmark & EIG + Warp & EIG & Sobol & Random & EIG + Warp & EIG & Sobol & Random \\ \hline
Cap ($S^2$)      & $\mathbf{16.9 \pm 1.6}$ & $-0.2 \pm 1.1$ & $-24.5 \pm 7.3$ & $-21.8 \pm 2.4$ & $\mathbf{30.8 \pm 1.3}$ & $-1.0 \pm 0.5$ & $-34.8 \pm 6.4$ & $-42.2 \pm 6.0$ \\
Two caps ($S^2$) & $\mathbf{8.4 \pm 1.6}$ & $-1.7 \pm 1.1$ & $-18.1 \pm 4.9$ & $-26.3 \pm 3.5$ & $\mathbf{16.1 \pm 0.9}$ & $-0.2 \pm 0.5$ & $-30.9 \pm 3.9$ & $-43.4 \pm 3.8$ \\
Cap ($S^3$)      & $\mathbf{2.9 \pm 1.2}$ & $1.4 \pm 0.9$ & $-9.2 \pm 3.1$ & $-7.8 \pm 2.9$ & $\mathbf{12.4 \pm 1.5}$ & $0.3 \pm 0.6$ & $-9.7 \pm 2.0$ & $-17.2 \pm 3.9$ \\
Two caps ($S^3$) & $\mathbf{0.5 \pm 2.1}$ & $-0.7 \pm 0.4$ & $-9.3 \pm 2.6$ & $-10.5 \pm 2.4$ & $\mathbf{6.7 \pm 0.9}$ & $0.0 \pm 0.6$ & $-10.0 \pm 1.8$ & $-18.8 \pm 2.8$ \\
\hline
\end{tabular}
\caption{Area reduction ($\mathrm{AR}$) relative to FPS under the $1$-NN evaluator on the sphere, using a $16{,}384$-point Sobol test set.
Results are the mean $\pm$ SEM over $10$ paired seeds (Appendix~\ref{app:sphere}); higher values are better.
The best result for each benchmark at each reported round is shown in bold.}
\label{tab:sphere}
\end{table}

We first examine covariance-based acquisition on the sphere. In \Cref{fig:sphere_warp}(b)-(e), we see that FPS and EIG achieve similar performance, and FPS will be used as the baseline.
EIG and FPS outperform Sobol and random sampling at both $20$ and $100$ rounds (\Cref{tab:sphere}), showing that sequentially adapting to the existing design is useful even without using the observed values.
Yet, EIG remains within $\pm 1.7\%$ of FPS across all settings.
With fixed hyperparameters, EIG is exactly open-loop by \Cref{prop:y_independence}: it depends on previous measurement locations, but not on their observed values.
We test whether this close agreement with FPS persists across covariance criteria and lengthscales.

\begin{figure}[t]
    \centering
    \includegraphics[width=0.6\linewidth]{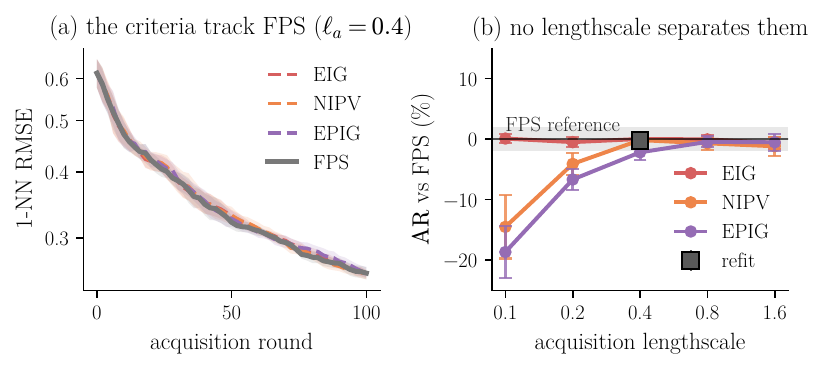}
    \caption{
    (a) At the default acquisition lengthscale, the covariance-based criteria closely track FPS throughout acquisition.
    (b) Varying the acquisition lengthscale does not improve on FPS. The light gray band marks $\pm 2\%$, and the $y$-axis matches that used for the cube in \Cref{fig:cube_ell}.}
    \label{fig:sphere_panel}
\end{figure}

\Cref{fig:sphere_panel}a shows that, at the default lengthscale, EIG, NIPV, and EPIG closely track FPS throughout acquisition.
\Cref{fig:sphere_panel}b tests whether this result depends on the acquisition lengthscale.
Across a sixteen-fold lengthscale sweep, EIG remains close to FPS, while NIPV and EPIG do not outperform FPS and deteriorate at shorter lengthscales.
Refitting the lengthscale by MAP at each round does not lead to improvement, because the fitted values remain in a range where performance is largely unchanged (\Cref{fig:sphere_panel}b).
Refitting introduces dependence on the observed values through a global lengthscale, but cannot represent spatial variation in the required resolution.

The reconstruction warp behaves differently.
On a sphere, the reconstruction warp realizes the target density by reparameterizing angular distance; this is the spherical counterpart of the transport map used on a cube.
The selected points concentrate around the transition band, where the target varies most rapidly (\Cref{fig:sphere_warp}).

On $S^2$, the reconstruction warp outperforms FPS on all $10$ seeds at both $20$ and $100$ rounds (\Cref{tab:sphere}), with larger gains at $100$ rounds.
The gains emerge later on $S^3$: AR is only $0.5$--$2.9\%$ after $20$ rounds, but increases to $6.7$--$12.4\%$ after $100$ rounds, when the warp outperforms FPS on all $10$ seeds for both targets.
This delayed improvement on $S^3$ is consistent with the posterior mean needing enough observations to resolve the relevant variation before the warp can use it effectively.

In sum, on a sphere, covariance-based acquisition offers little improvement over FPS, whereas the reconstruction warp can improve measurement allocation by using posterior-mean variation. Inherently boundary-free experimental domains are relatively uncommon, but do arise, for example, in sensor placement on spherical surfaces and in design problems on Lie groups.\looseness-1

\subsection{On bounded domains}
\label{sec:cube}
\label{sec:cube_flip}

\begin{figure}[t]
    \centering
    \includegraphics[width=\linewidth]{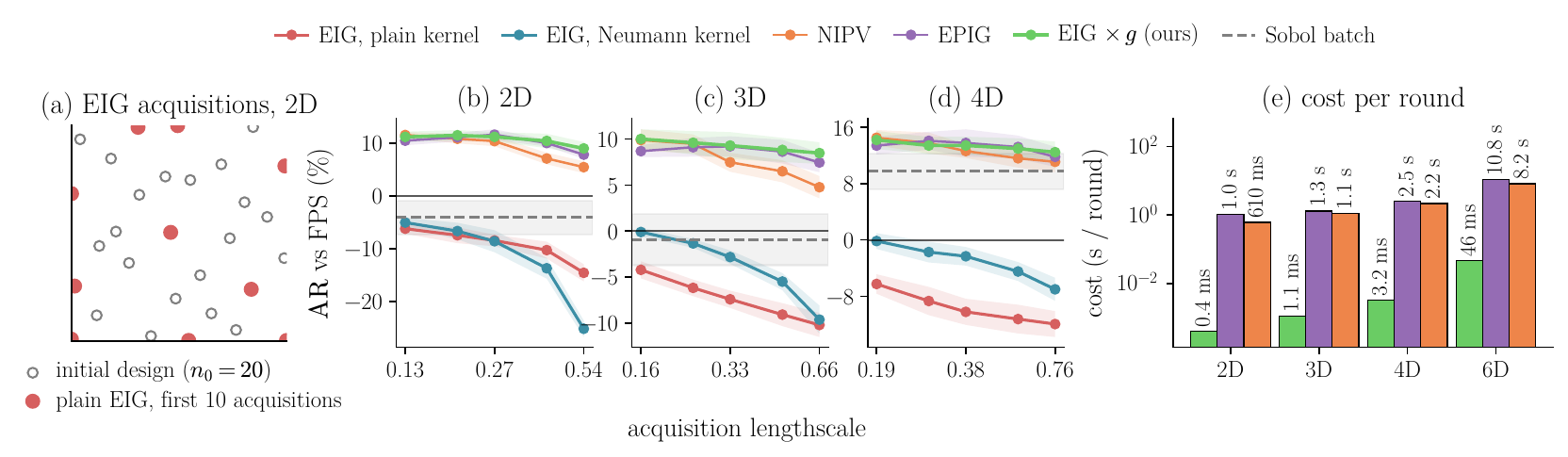}
    \caption{
    (a) Twenty points acquired by plain EIG in 2D, showing its concentration near the boundary.
    (b--d) Area reduction ($\mathrm{AR}$) relative to FPS over $100$ rounds across acquisition lengthscales, grouped by dimension. The geometric equalizer and the integrated criteria NIPV and EPIG remain above FPS across all tested lengthscales and dimensions.
    (e) Per-round acquisition cost on a logarithmic scale. The precomputed geometric equalizer is inexpensive to evaluate, whereas NIPV and EPIG require numerical integration at every round.\looseness-1}
    \label{fig:cube_ell}
\end{figure}

We now restore the boundary and consider a cube $[0,1]^D$.
Unlike on a sphere, EIG can now perform worse than a fixed space-filling design. Across the lengthscales tested in \Cref{fig:cube_ell}b--d, EIG falls below Sobol on four functions by $1.8$--$12.6$ percentage points in AR, while remaining close to FPS. Since EIG closely tracks FPS on a sphere, the poorer performance on a cube points to boundary bias as the cause.\looseness-1


The effect is visible directly in the selected experiment designs.
In two dimensions, EIG places $56$--$58\%$ of the acquired points within $0.05$ of a face, increasing to $84$--$99\%$ in 3D, compared with $20\%$ and $27\%$ for Sobol (\Cref{fig:cube_ell}a; Appendix~\ref{app:selection_bias}).
Near a corner, only about $2^{-D}$ of a symmetric local neighborhood lies inside the domain, while the truncated correlation neighborhood also gives these locations higher posterior variance.
Random sampling performs worse than Sobol, indicating that Sobol's advantage comes from its space-filling geometry rather than simply from being fixed in advance.\looseness-1

Applying the geometric equalizer changes this behavior.
Across all tested dimensions and acquisition lengthscales, EIG corrected with the equalizer achieves $8$--$14\%$ AR relative to FPS, whereas plain EIG does not consistently outperform the fixed Sobol design (\Cref{fig:cube_ell}b--d).
NIPV and EPIG also overcome the boundary bias but require a new domain integral at every round; the precomputed geometric equalizer is two to three orders of magnitude cheaper to evaluate (\Cref{fig:cube_ell}e).
\Cref{fig:main_curves} shows representative reconstruction-error trajectories, with all benchmarks reported in Appendix~\ref{app:all_curves}.
On Levy 4D and Styblinski--Tang 3D, the error under plain EIG even increases during acquisition, whereas every boundary-corrected method continues to improve.
\looseness=-1

With boundary bias corrected, we can now ask what observation dependence adds.
\Cref{tab:main_results} compares the geometric equalizer alone, the equalizer combined with the reconstruction warp, and the equalizer combined with a trained warp.
The geometric equalizer alone improves on FPS on every benchmark (\Cref{tab:main_results,fig:main_curves}), despite remaining observation-independent.
At $200$ rounds, the full method matches or improves on the equalizer across all benchmarks; at $100$ rounds, its mean AR is higher on all except Levy 4D.
The largest gain occurs on Gramacy--Lee, where the full method reaches $44.5\%$ AR at $200$ rounds, $37.1$ percentage points above the equalizer alone.\looseness-1

\begin{table}[t]
\centering
\footnotesize
\setlength{\tabcolsep}{2pt}
\resizebox{\linewidth}{!}{%
\begin{tabular}{lccc|ccc|ccc}
\toprule
& \multicolumn{3}{c}{50 rounds}
& \multicolumn{3}{c}{100 rounds}
& \multicolumn{3}{c}{200 rounds} \\
\cmidrule(lr){2-4}
\cmidrule(lr){5-7}
\cmidrule(lr){8-10}

Function
& \shortstack{$g$ only}
& \shortstack{$g$ + recon.\\warp}
& \shortstack{$g$ + Kumar.\\warp, trained}
& \shortstack{$g$ only}
& \shortstack{$g$ + recon.\\warp}
& \shortstack{$g$ + Kumar.\\warp, trained}
& \shortstack{$g$ only}
& \shortstack{$g$ + recon.\\warp}
& \shortstack{$g$ + Kumar.\\warp, trained} \\
\midrule
Peaks (2D)             & $12.0 \pm 0.8$ & $\mathbf{20.4} \pm 1.3$ & $17.4 \pm 0.9$ & $10.4 \pm 0.5$ & $\mathbf{21.7} \pm 1.0$ & $18.1 \pm 0.7$ & $9.3 \pm 0.4$ & $\mathbf{22.8} \pm 0.5$ & $18.4 \pm 0.5$ \\
Smooth box (2D)        & $1.6 \pm 2.2$ & $\mathbf{21.4} \pm 1.9$ & $8.7 \pm 3.1$ & $1.1 \pm 2.7$ & $\mathbf{27.6} \pm 1.2$ & $12.9 \pm 2.4$ & $1.9 \pm 2.2$ & $\mathbf{34.3} \pm 1.1$ & $15.5 \pm 2.2$ \\
Gramacy--Lee (2D)      & $9.1 \pm 1.5$ & $\mathbf{37.1} \pm 1.8$ & $29.0 \pm 1.9$ & $9.3 \pm 1.3$ & $\mathbf{41.7} \pm 1.5$ & $32.4 \pm 1.4$ & $7.4 \pm 1.0$ & $\mathbf{44.5} \pm 1.1$ & $33.9 \pm 1.1$ \\
Levy (2D)              & $17.1 \pm 1.8$ & $\mathbf{25.5} \pm 1.5$ & $18.7 \pm 2.2$ & $13.1 \pm 1.3$ & $\mathbf{25.9} \pm 1.1$ & $13.4 \pm 1.6$ & $9.7 \pm 1.0$ & $\mathbf{25.5} \pm 0.8$ & $8.6 \pm 1.5$ \\
Zhou98 (2D)            & $8.2 \pm 0.7$ & $\mathbf{20.0} \pm 1.5$ & $14.5 \pm 1.3$ & $8.4 \pm 0.6$ & $\mathbf{23.2} \pm 1.0$ & $16.3 \pm 1.0$ & $7.3 \pm 0.5$ & $\mathbf{24.9} \pm 0.8$ & $16.8 \pm 0.7$ \\
Beale (2D)             & $25.4 \pm 2.2$ & $\mathbf{38.2} \pm 1.7$ & $16.2 \pm 2.6$ & $22.3 \pm 1.7$ & $\mathbf{40.4} \pm 1.4$ & $19.0 \pm 1.9$ & $18.9 \pm 1.5$ & $\mathbf{41.6} \pm 1.1$ & $21.2 \pm 1.5$ \\
Smooth box (3D)        & $8.7 \pm 2.5$ & $\mathbf{12.8} \pm 2.0$ & $10.3 \pm 2.2$ & $10.0 \pm 2.1$ & $\mathbf{14.4} \pm 2.3$ & $\mathbf{13.5} \pm 1.5$ & $10.9 \pm 1.4$ & $\mathbf{18.3} \pm 1.4$ & $15.4 \pm 1.3$ \\
Hartmann (3D)          & $9.2 \pm 0.7$ & $\mathbf{13.2} \pm 0.7$ & $5.3 \pm 1.3$ & $10.3 \pm 0.6$ & $\mathbf{16.0} \pm 0.5$ & $6.1 \pm 0.9$ & $9.9 \pm 0.5$ & $\mathbf{17.1} \pm 0.4$ & $4.8 \pm 0.7$ \\
Levy (3D)              & $19.3 \pm 1.2$ & $\mathbf{20.5} \pm 1.3$ & $18.5 \pm 1.2$ & $\mathbf{22.0} \pm 0.9$ & $\mathbf{22.8} \pm 1.2$ & $20.6 \pm 1.4$ & $21.0 \pm 0.9$ & $\mathbf{22.4} \pm 1.3$ & $18.9 \pm 1.5$ \\
Styblinski--Tang (3D)  & $24.8 \pm 1.2$ & $\mathbf{25.9} \pm 0.9$ & $24.2 \pm 1.0$ & $25.5 \pm 1.0$ & $\mathbf{26.7} \pm 0.9$ & $23.2 \pm 1.1$ & $23.4 \pm 0.6$ & $\mathbf{25.4} \pm 0.8$ & $21.1 \pm 1.0$ \\
Smooth box (4D)        & $7.4 \pm 2.2$ & $\mathbf{12.8} \pm 2.2$ & $8.7 \pm 2.1$ & $9.2 \pm 2.1$ & $\mathbf{14.2} \pm 2.4$ & $11.5 \pm 1.9$ & $7.8 \pm 1.6$ & $\mathbf{12.3} \pm 1.8$ & $\mathbf{12.4} \pm 1.5$ \\
Hartmann (4D)          & $8.0 \pm 0.6$ & $\mathbf{8.7} \pm 0.5$ & $6.5 \pm 0.6$ & $9.9 \pm 0.6$ & $\mathbf{10.9} \pm 0.5$ & $7.6 \pm 0.7$ & $10.2 \pm 0.4$ & $\mathbf{11.8} \pm 0.4$ & $7.2 \pm 0.6$ \\
Levy (4D)              & $\mathbf{18.5} \pm 1.0$ & $\mathbf{18.1} \pm 1.3$ & $\mathbf{17.5} \pm 1.4$ & $\mathbf{20.7} \pm 1.2$ & $\mathbf{20.6} \pm 1.4$ & $19.1 \pm 1.3$ & $\mathbf{20.8} \pm 1.3$ & $\mathbf{21.1} \pm 1.5$ & $18.7 \pm 1.2$ \\
Power plant (4D, real) & $3.2 \pm 1.1$ & $\mathbf{5.5} \pm 1.0$ & $2.9 \pm 1.1$ & $3.5 \pm 0.9$ & $\mathbf{5.9} \pm 1.0$ & $2.2 \pm 1.4$ & $2.6 \pm 0.9$ & $\mathbf{6.1} \pm 0.8$ & $0.9 \pm 1.4$ \\
Friedman (5D)          & $\mathbf{9.1} \pm 0.5$ & $\mathbf{9.4} \pm 0.4$ & $2.9 \pm 0.7$ & $11.3 \pm 0.4$ & $\mathbf{11.8} \pm 0.2$ & $3.5 \pm 0.6$ & $12.5 \pm 0.4$ & $\mathbf{13.2} \pm 0.2$ & $3.6 \pm 0.5$ \\
Levy (5D)              & $17.7 \pm 1.0$ & $\mathbf{18.9} \pm 1.0$ & $15.5 \pm 0.9$ & $21.8 \pm 1.0$ & $\mathbf{23.0} \pm 1.0$ & $19.0 \pm 0.9$ & $23.8 \pm 1.0$ & $\mathbf{25.2} \pm 1.1$ & $21.0 \pm 1.0$ \\
Airfoil (5D, real)     & $\mathbf{7.0} \pm 1.0$ & $\mathbf{7.3} \pm 0.9$ & $2.1 \pm 0.8$ & $\mathbf{8.9} \pm 1.4$ & $\mathbf{9.6} \pm 1.2$ & $3.6 \pm 1.1$ & $\mathbf{11.1} \pm 1.4$ & $\mathbf{11.6} \pm 1.2$ & $5.7 \pm 1.2$ \\
Hartmann (6D)          & $2.2 \pm 1.7$ & $\mathbf{3.7} \pm 1.5$ & $1.8 \pm 1.2$ & $\mathbf{5.0} \pm 1.3$ & $\mathbf{5.6} \pm 1.2$ & $4.0 \pm 0.9$ & $\mathbf{6.7} \pm 1.0$ & $\mathbf{6.8} \pm 1.2$ & $\mathbf{6.0} \pm 0.8$ \\
\bottomrule
\end{tabular}}
\caption{
Area reduction ($\mathrm{AR}$, \%) relative to FPS after 50, 100, and 200 acquisition rounds
(mean $\pm$ SEM over 10 seeds; higher is better).
Here, $g$ denotes the geometric equalizer; $g$ + recon. warp adds the reconstruction warp, while $g$ + Kumar. warp uses the trained Kumaraswamy warp instead.
At each round count, the best result for each function is shown in bold together
with results within one standard error.
At 200 rounds, the reconstruction warp matches or exceeds the equalizer alone
on every benchmark, whereas the trained warp underperforms the equalizer on
11 of the 18 benchmarks.
Rows marked \emph{real} are real-data benchmarks evaluated against a GP surrogate
fitted to the full dataset (\Cref{app:real_test_functions}).
}
\label{tab:main_results}
\end{table}

\begin{figure}[t]
    \centering
    \includegraphics[width=\linewidth]{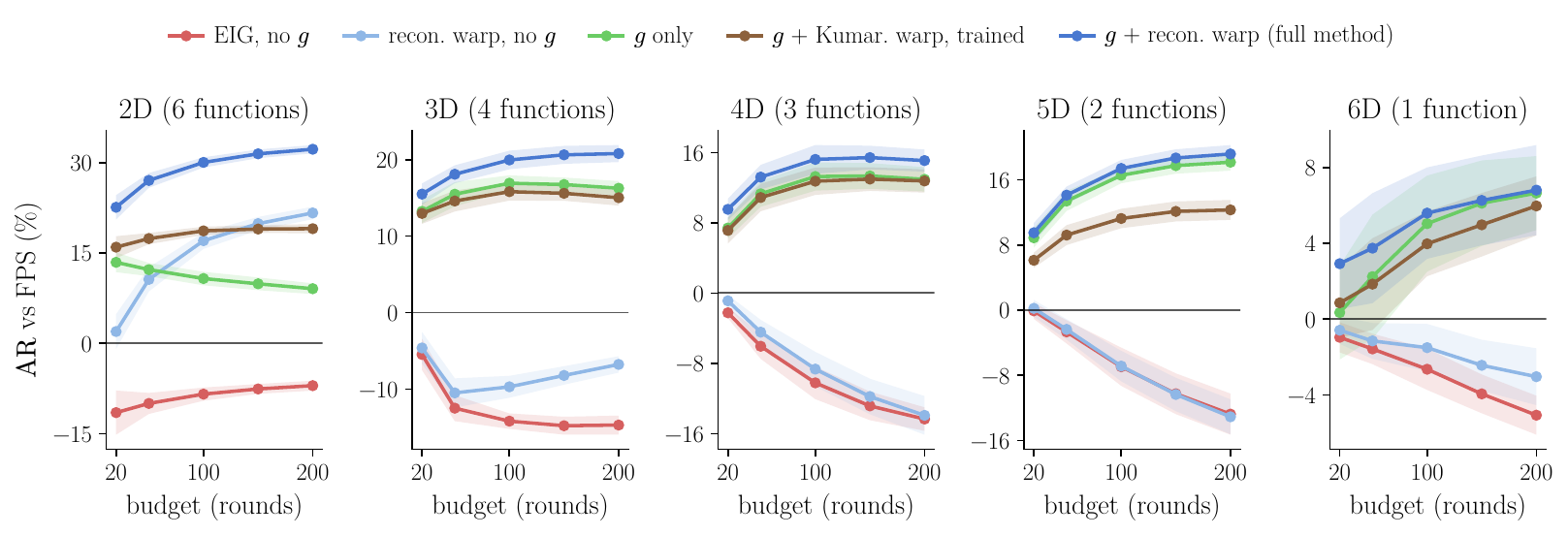}
    \caption{
    Effect of the geometric equalizer and reconstruction warp, separately and in combination, as acquisition progresses.
    Area reduction ($\mathrm{AR}$) relative to FPS is shown by dimension; lines give the mean over $10$ seeds and shaded regions show $95\%$ confidence intervals.
    Without the geometric equalizer, the reconstruction warp improves on FPS only in 2D.
    With boundary bias corrected, the equalizer provides most of the early gain, while the contribution of the reconstruction warp becomes larger over later rounds.}
    \label{fig:ar_budget}
\end{figure}

The warp's additional contribution generally decreases with dimension and grows as acquisition progresses (\Cref{fig:ar_budget}).
Hartmann 6D is the clearest case in which the warp adds little beyond the equalizer. After $200$ rounds, the warped and unwarped results are nearly identical, consistent with the posterior mean not yet resolving enough local variation to guide the warp.
The same dimensional trend appears on the boundary-free sphere: at $100$ rounds, the warp gain falls from $30.8\%$ to $12.4\%$ between the $S^2$ and $S^3$ single-cap targets (\Cref{fig:sphere_warp}).
On the smooth-box benchmarks, the additional gain over the equalizer falls from $26.5$ percentage points in 2D to $4.4$ in 3D.
Since the trend also appears without a boundary, it cannot be attributed solely to boundary bias.
Instead, it is consistent with higher-dimensional designs remaining sparser for a fixed number of acquisition rounds. This makes it harder for the posterior mean to resolve the local function structure needed to define an effective warp.\looseness-1

We also compare against a trainable Kumaraswamy warp, refitted by marginal likelihood at every round under the same geometric equalizer (\Cref{app:warp_details}).
At $200$ rounds, the trained warp underperforms the equalizer alone on nine of the sixteen synthetic benchmarks, including an $8.9$-percentage-point loss on Friedman (\Cref{tab:main_results}).
Hartmann 3D and Levy 4D illustrate similar failures in the reconstruction trajectories (\Cref{fig:main_curves}).
Unlike the trained warp, the reconstruction warp retains an explicit uniform component through $\rho_0$, which limits how strongly the acquisition geometry can deform when the posterior-mean gradients provide little useful spatial information. This helps prevent the warp from degrading performance. \looseness-1

We also evaluate the reconstruction warp without the geometric equalizer (\Cref{fig:ar_budget}).
In 2D it outperforms FPS on all six functions, indicating that the observation-dependent allocation is strong enough to overcome boundary bias in these cases.
In higher dimensions, however, AR becomes negative on nine of ten benchmarks and approaches plain EIG performance.

\begin{figure}[t]
    \centering
    \includegraphics[width=0.9\linewidth]{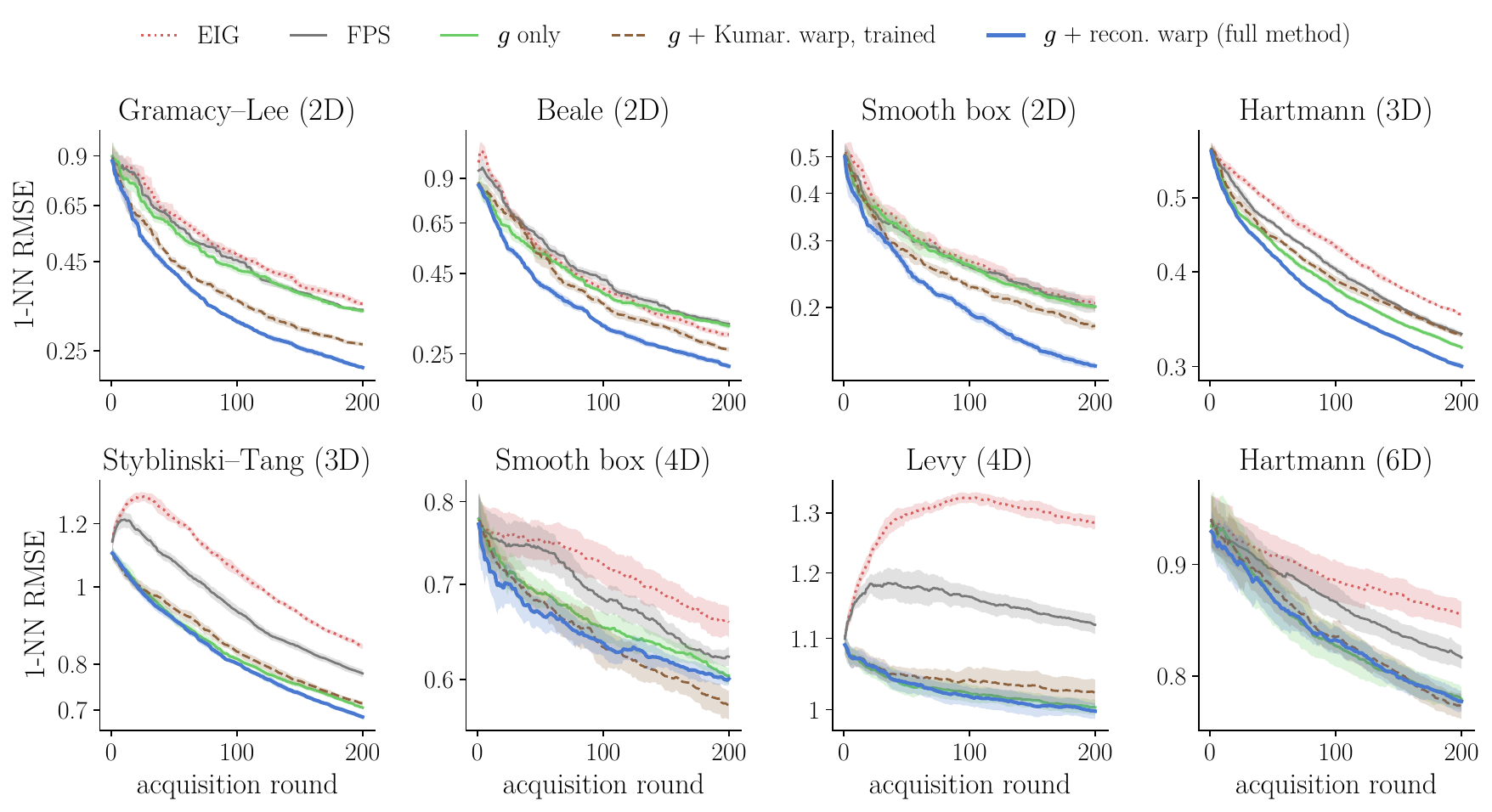}
    \caption{
    Reconstruction error over acquisition rounds for eight of the sixteen synthetic benchmarks in two to six dimensions.
    Curves show the mean over $10$ seeds, with shaded regions indicating $95\%$ confidence intervals.
    On Levy 4D and Styblinski--Tang 3D, the error under plain EIG increases during acquisition, whereas every equalized method continues to improve.}
    \label{fig:main_curves}
\end{figure}

In addition to synthetic functions, we evaluate two real-data benchmarks from the UCI repository: combined-cycle power-plant output (4D)~\citep{combined_cycle_power_plant_294} and airfoil self-noise (5D)~\citep{airfoil_self-noise_291}, marked \emph{real} in \Cref{tab:main_results}.
They are scored against GP surrogates fitted to the full datasets.
These surrogates are used only for evaluation and never for acquisition (Appendix~\ref{app:real_test_functions}).
The same qualitative pattern holds: plain EIG performs worse than FPS, the geometric equalizer restores an advantage over FPS, and adding the reconstruction warp gives the highest mean AR at each reported round count.
The warp advantage grows with iterations on the power-plant benchmark but remains smaller on airfoil.
The real-data benchmarks show the same behavior; the equalizer provides a consistent improvement, while the warp contributes more when the posterior mean resolves enough variation to guide acquisition. \looseness-1

\section{Discussion}
\label{sec:discussion}
The experiments show that boundary bias and observation independence affect sequential design in distinct ways.
On a boundary-free domain, EIG, NIPV, and EPIG closely track FPS, and refitting hyperparameters changes little.
On a cube, boundary bias can cause a Sobol design (fixed before any values are observed) to \textit{outperform} sequential variance maximization early in acquisition.
Together, these results show that adapting to where points have been placed is useful, but relying on covariance alone is not sufficient to determine where the observed function requires additional resolution. The geometric equalizer corrects boundary bias, while the reconstruction warp uses the posterior mean to redirect measurements toward regions that need finer resolution.

The two corrections contribute differently over the course of acquisition. The geometric equalizer provides most of the early improvement, whereas the contribution of the reconstruction warp grows as the posterior mean resolves more local structure. The additional benefit of the reconstruction warp also decreases with dimension. For a fixed budget, higher-dimensional designs are sparser, so the posterior mean resolves local function variation less accurately and provides weaker guidance for the warp.

Using the same fixed reconstruction rule to evaluate all acquired designs provides a fair comparison between acquisition strategies, because differences in performance then reflect the quality of the acquired designs rather than differences in the reconstruction model used for evaluation.

Applying our method in real experiment settings has huge potential to reduce experimental costs. Overcoming both described problems will prevent oversampling of the boundary regions, and make sequential experiment design more efficient. \looseness-1

\paragraph{Limitations and future work.}
Deriving the reconstruction-optimal density, we approximate the unknown function locally by its first-order linear expansion around each sample point, which may not hold for functions with large local curvature.
The reconstruction density also depends on the accuracy of the posterior mean early in acquisition. This could lead to unexplored regions never being sampled, because they revert to the posterior mean and appear to be flat. The uniform component ensures a minimum sampling rate, but does not change the gradient estimation.

The main experiments use a fixed acquisition kernel and pool-based active learning.
Extending the full method to jointly adapt the acquisition hyperparameters or to use gradient-based acquisition optimization remains open. The current Knothe--Rosenblatt construction is constrained by coordinate ordering. Parameterizing the transport map via more expressive flow-based methods, such as continuous normalizing flows or neural ODE-based maps, or rotation-invariant optimal-transport maps could capture more complex geometry. Finally, our experiments focus on global reconstruction.
Whether the same separation between geometric correction and observation-dependent allocation is useful for optimization remains an open question.

\section{Conclusion}
Sequential experiment design with standard GPs and variance-based acquisition has two shortcomings: boundary bias and observation independence, where the acquisition depends on where measurements were taken but not on their observed values. We developed a two-step correction framework that separates acquisition and prediction. First, a geometric equalizer corrects boundary bias without using observations. Then, a reconstruction warp uses variation in the posterior mean to allocate measurements where finer resolution is needed. The geometric equalizer efficiently removes boundary bias and improves performance across all evaluated functions. When the posterior mean can resolve enough local structure, the warp further improves performance. Together, these results show that boundary bias and observation independence are distinct limitations of naive variance-based sequential experiment design and should be addressed separately. Consequently, our framework can significantly reduce experimental cost in practical applications.


\bibliography{references}
\bibliographystyle{tmlr}

\appendix

\section{Derivation of EIG for Gaussian processes}
\label{app:eig_derivation}

For a candidate location $\vx_*$, the expected information gain (EIG) is the
mutual information between the noisy observation $y_*$ and the latent function,
conditioned on the current data:
\begin{equation}
    \mathrm{EIG}(\vx_*)
    =
    I(y_*;z\mid\mathcal D,\vx_*)
    =
    H[y_*\mid\mathcal D,\vx_*]
    -
    H[y_*\mid\mathcal D,\vx_*,z].
    \label{eq:eig_mi}
\end{equation}

Under the GP posterior,
\begin{equation}
    y_*\mid\mathcal D,\vx_*
    \sim
    \mathcal N\!\bigl(
        \mu(\vx_*),
        s^2(\vx_*)+\sigma_n^2
    \bigr),
\end{equation}
whereas conditioning on the latent function leaves only the observation noise,
$y_*\mid z\sim\mathcal N(z(\vx_*),\sigma_n^2)$.
Using the entropy of a univariate Gaussian therefore gives
\begin{align}
    \mathrm{EIG}(\vx_*)
    &=
    \frac12
    \log
    \left(
        \frac{s^2(\vx_*)+\sigma_n^2}
             {\sigma_n^2}
    \right),
\end{align}
which is \eqref{eq:eig_gp}.

Because $s^2(\vx_*)$ depends on the observation locations $X$ and kernel
hyperparameters $\theta$, but not on the observed values $\vy$, EIG is
observation-independent under fixed hyperparameters, as stated in
\Cref{prop:y_independence}.

\subsection{Alternative covariance-based criteria}
\label{app:zoo_criteria}

We also compare against negative integrated posterior variance (NIPV) and
expected predictive information gain (EPIG). Under fixed GP hyperparameters,
both depend only on posterior covariances and are therefore
observation-independent.

\paragraph{Negative integrated posterior variance.}
For a candidate $\vx_*$, the reduction in posterior variance at a target
location $\vx^u$ is
\begin{equation}
    \frac{
        \mathrm{Cov}\!\bigl(z(\vx^u),z(\vx_*)\bigr)^2
    }{
        s^2(\vx_*)+\sigma_n^2
    }.
\end{equation}
We therefore use
\begin{equation}
    \mathrm{NIPV}(\vx_*)
    =
    \int
    \frac{
        \mathrm{Cov}\!\bigl(z(\vx^u),z(\vx_*)\bigr)^2
    }{
        s^2(\vx_*)+\sigma_n^2
    }
    p^*(\vx^u)\,d\vx^u,
\end{equation}
estimated numerically over a low-discrepancy target set.

\paragraph{Expected predictive information gain.}
EPIG~\citep{smith2023prediction} measures
\begin{equation}
    \mathrm{EPIG}(\vx_*)
    =
    I(y_*;\vy^*
      \mid
      \mathcal D,\vx_*,X^*),
    \label{eq:epig_mi}
\end{equation}
where $\vy^*$ denotes noisy predictions at target locations $X^*$.

Under the GP posterior,
\begin{equation}
\begin{pmatrix}
    y_*\\
    \vy^*
\end{pmatrix}
\Bigg|\mathcal D
\sim
\mathcal N
\left(
\begin{pmatrix}
    \mu(\vx_*)\\
    \vmu^*
\end{pmatrix},
\begin{pmatrix}
    s^2(\vx_*)+\sigma_n^2
        & \vs_*^{*\top}\\
    \vs_*^*
        & S^{**}+\sigma_n^2I
\end{pmatrix}
\right).
\end{equation}
The Gaussian mutual information reduces to
\begin{equation}
    \mathrm{EPIG}(\vx_*)
    =
    \frac12
    \log
    \left(
    \frac{
        s^2(\vx_*)+\sigma_n^2
    }{
        s^2(\vx_*)+\sigma_n^2
        -
        \vs_*^{*\top}
        (S^{**}+\sigma_n^2I)^{-1}
        \vs_*^*
    }
    \right).
    \label{eq:epig_gp}
\end{equation}

EPIG therefore has a richer geometric dependence than pointwise EIG:
its score depends on the spatial relationship between the candidate, the
target locations, and the existing observations. Under fixed hyperparameters,
however, these covariance quantities remain independent of the observed
values.

\section{Exact geometric neutrality}
\label{app:neutrality}

The geometric equalizer of Appendix \ref{sec:boundary_calibration} is a clipped
practical realization of an exact neutrality construction.

Let
\[
    \bar v_N(\vx)
    =
    \mathbb E_{X_N\sim Q_N}
    [s^2(\vx\mid X_N)]
\]
be the reference variance from \eqref{eq:reference_variance}.
For a pointwise multiplicative weighting $w_N(\vx)$, geometric neutrality
requires the expected acquisition score under the reference designs to be
constant across the domain:
\begin{equation}
    \mathbb E_{X_N\sim Q_N}
    \left[
        w_N(\vx)s^2(\vx\mid X_N)
    \right]
    =
    w_N(\vx)\bar v_N(\vx)
    =
    C.
    \label{eq:flat_expected_score}
\end{equation}
Hence
\begin{equation}
    w_N^*(\vx)
    =
    \frac{C}{\bar v_N(\vx)}.
    \label{eq:unique_normalizer}
\end{equation}
Within the class of pointwise multiplicative weights satisfying
\eqref{eq:flat_expected_score}, this correction is unique up to the global
scale $C$, which does not affect the acquisition argmax.

The practical equalizer in \eqref{eq:geo_equalizer} uses
$C=\bar v_{\mathrm{int}}$ and clips the ratio at one.
The clipping suppresses numerical fluctuations in the nearly flat interior
and guarantees $g(\vx)\leq1$, so the correction never amplifies the original
acquisition score.

\section{Selection-bias diagnostic}
\label{app:selection_bias}

We use the function-free diagnostic of \citet{bankestad2026boundary} to
measure the spatial preference induced by each acquisition criterion.
Candidates are stratified by distance to the boundary, subsampled according
to the corresponding shell volumes, and the acquisition argmax is recorded
over independent scrambled-Sobol designs.
No function values enter the diagnostic.

\begin{figure}[ht]
    \centering
    \includegraphics[width=\linewidth]{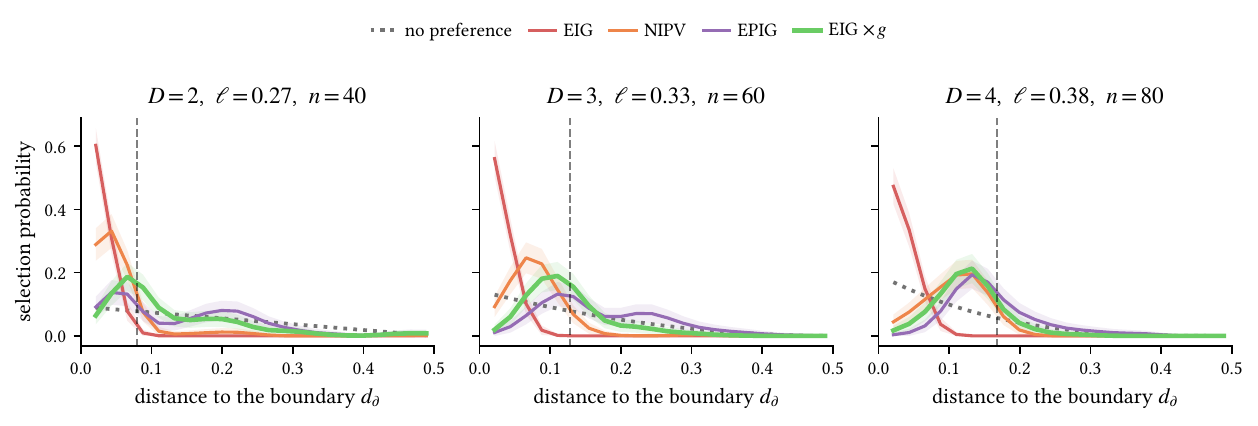}
    \caption{
    Selection profiles induced by the acquisition criteria before any
    function values are observed, in two to four dimensions.
    Variance maximization concentrates near the boundary; NIPV and EPIG
    shift selections inward but retain preferred shells, while the geometric
    equalizer moves the mode toward the interior and produces a profile
    close to EPIG.
    The dotted line denotes spatially unbiased random selection and is shown
    as a reference rather than as a target design.}
    \label{fig:selection_bias}
\end{figure}

Plain variance maximization places most selections close to the boundary.
NIPV and EPIG, which score candidates through their effect over the domain,
shift the selections inward.
The equalized criterion, EIG$\times g$, moves the mode farther from the
boundary and substantially reduces the mass in the boundary shell.

The expected-score neutrality of Appendix \ref{app:neutrality} should not be confused
with a uniform distribution of selected argmax locations.
The former concerns the mean acquisition score under reference designs;
the latter also depends on spatial fluctuations around that mean.

\section{Reconstruction-warp implementation} \label{app:warp_details} 
\subsection{Numerical transport}

We evaluate $\nabla\mu_{\mathrm a}$ on scrambled-Sobol probes and use these values to construct the target density in \eqref{eq:target_density}. We use $8192$ probes for $D\leq3$ and $16{,}384$ for $D\geq4$. The normalization constant $c$ is the root-mean-square probe-gradient magnitude, with a numerical floor of $\varepsilon=10^{-8}$.

The Knothe--Rosenblatt map is constructed from numerical marginal and
conditional CDFs.
Conditional CDFs are estimated using histogram bins within conditioning
cells and interpolated between neighboring cell centers.
The resulting coordinate maps are monotone, piecewise linear, and pinned at
the endpoints, giving a bijection of $[0,1]^D$.
The map is rebuilt directly from the posterior mean at every acquisition
round and requires no optimization.

\subsection{Trainable Kumaraswamy baseline}

The trained-warp baseline uses the Kumaraswamy input warp
\citep{kumar_warp},
\begin{equation}
    T_\phi(x)
    =
    1-(1-x^a)^b,
    \qquad
    \phi=\{a,b\},
    \label{eq:kumar}
\end{equation}
applied independently to each input dimension.

The warp modifies only the acquisition GP, as does the reconstruction warp;
the prediction GP remains unchanged.
At each acquisition round, the parameters are initialized at the identity
$a=b=1$ and refitted by maximizing the exact marginal likelihood of the
acquisition GP.
Kernel lengthscales and observation noise remain fixed to the values of the
experimental protocol.
The parameters are optimized in a bounded unconstrained parameterization
corresponding to $(a,b)\in(0.2,3)$ using Adam for $100$ steps.

The trained arm in \Cref{tab:main_results} uses the same geometric equalizer
as the full method, so the comparison isolates the choice of warp.

\section{Boundary-free sphere experiment}
\label{app:sphere}

The spherical experiment of \Cref{sec:baselines_reveal} removes boundary
effects while preserving the reconstruction problem.

\paragraph{Protocol.}
Each benchmark uses ten paired seeds and $100$ acquisition rounds.
The initial designs contain $20$ points on $S^2$ and $40$ points on $S^3$.
Area reduction is computed per seed relative to FPS started from the same
initial design, following Appendix \ref{app:area_reduction}.

\paragraph{Targets.}
The cap targets have the form
\[
    f(\vx)
    =
    \sigma
    \left(
        k(\langle\vx,\vc\rangle-\cos\psi_0)
    \right),
\]
with center direction $\vc$, angular radius $\psi_0$, and steepness $k$.
The two-cap targets combine two such regions.

\paragraph{Kernel.}
The GP uses the chordal Mat\'ern-$5/2$ kernel
\[
    k(\vx,\vx')
    =
    \kappa(\|\vx-\vx'\|),
\]
which is positive definite on the sphere
\citep{gneiting2013strictly}.
The acquisition kernel is fixed throughout a run and shared across methods.

The lengthscales follow the same design-spacing rule used on the cube.
The values used in the experiments lie within the broad plateau shown in
\Cref{fig:sphere_panel}; per-round MAP refitting remains within this regime,
which explains its limited effect on acquisition performance.

\paragraph{Candidate and integration sets.}
Acquisition is performed over a dense deterministic near-uniform candidate
set.
NIPV and EPIG use low-discrepancy integration sets dense relative to the
shortest acquisition lengthscale.
All reported experiments operate well above the discretization scale of
these sets.

\paragraph{Design density.}
On $S^d$ we use
\begin{equation}
    \hat\rho(\vx)
    =
    1
    +
    \left(
        \frac{
            \|\nabla_{\!\top}\mu(\vx)\|
        }{c}
    \right)^{\frac{2d}{d+2}},
\end{equation}
where $\nabla_{\!\top}\mu$ is the posterior-mean gradient projected onto the
tangent space and $c$ is its root-mean-square magnitude, floored at
$\varepsilon=10^{-8}$.
The exponent uses the intrinsic dimension $d$.

\paragraph{Warp.}
For an axially symmetric density around a center direction $\vc$, let
\[
    \psi
    =
    \arccos\langle\vx,\vc\rangle
\]
be the colatitude.
Let $F(\psi)$ denote the cumulative distribution induced by $\hat\rho$ under
the spherical measure and
\[
    G(\psi)=\frac{1-\cos\psi}{2}
\]
the uniform cumulative.
The warped colatitude is
\[
    \psi'=G^{-1}(F(\psi)).
\]
The point is moved along the same great circle through $\vc$.
For the cap targets this realizes the same density-matching principle as the
Knothe--Rosenblatt construction on the cube.
The map is rebuilt every round without training and reduces to the identity
for an exactly flat posterior mean.

\paragraph{Low-discrepancy baseline.}
On $S^2$, the Sobol baseline is obtained by mapping a two-dimensional
scrambled Sobol sequence through the Lambert equal-area construction.
On $S^3$, scrambled Sobol points are transformed through the inverse-normal
CDF and normalized to unit length.
Both constructions produce approximately uniform low-discrepancy sequences
whose prefixes define the fixed baseline designs.

\paragraph{Alternative evaluator.}
As a check on the nearest-neighbor evaluator, we also evaluated the acquired
designs using the prediction-GP posterior mean.
The resulting rankings and qualitative conclusions agree with the primary
$1$-NN evaluator; the $1$-NN results are reported in the main text.

\section{Extended related work}
\label{app:extended_related_work}

The main related-work discussion focuses on the distinction most relevant to this paper: whether observation dependence is introduced through the predictive model or separately through acquisition.
\Cref{tab:related_work} summarizes representative approaches.

\begin{table}[ht]
    \centering
    \caption{
    Representative approaches for modifying spatial allocation in GP-based sequential design.
    ``Prediction modified'' indicates whether the mechanism also changes the
    model used for prediction.}
    \label{tab:related_work}
    \footnotesize
    \setlength{\tabcolsep}{3pt}
    \begin{tabular}{l c c l}
        \toprule
        Method
        & Prediction modified
        & How observations enter
        & How spatial allocation is changed \\
        \midrule

        Input warp~\citep{kumar_warp}
            & Yes
            & Via fitting
            & Learned input transformation \\

        DKL~\citep{wilson2016deep}
            & Yes
            & Via fitting
            & Learned representation \\

        Manifold GP~\citep{calandra2016manifold}
            & Yes
            & Via fitting
            & Learned input transformation \\

        Deep GP~\citep{damianou2013deep}
            & Yes
            & Via fitting
            & Learned nonstationarity \\

        Nonstationary GP~\citep{heinonen2016non}
            & Yes
            & Via fitting
            & Spatially varying covariance \\

        EPIG~\citep{smith2023prediction}
            & No
            & No
            & Integrated predictive influence \\

        \midrule

        \textbf{Ours}
            & \textbf{No}
            & \textbf{Via posterior mean}
            & \textbf{Reconst.-driven input warp} \\

        \bottomrule
    \end{tabular}
\end{table}

Learned input warps~\citep{kumar_warp,balandat2020botorch}, manifold
GPs~\citep{calandra2016manifold}, deep kernel learning
\citep{wilson2016deep}, deep GPs~\citep{damianou2013deep,sauer2023active},
and explicitly nonstationary kernels~\citep{heinonen2016non} allow
observations to reshape the covariance used for prediction and acquisition.
Our approach instead leaves the prediction GP unchanged and modifies only
the acquisition geometry.

Prediction-oriented criteria such as EPIG~\citep{smith2023prediction} score candidates by their influence on predictions across the domain. Other approaches modify the surrogate to change how acquisition responds to spatial structure~\citep{bodin2020modulating}. Our reconstruction warp instead leaves the prediction GP unchanged: its spatial allocation comes from the reconstruction objective and is realized directly in the acquisition geometry.\looseness-1

\section{Experimental configurations}
\label{app:configurations}

\paragraph{Common protocol.}
All bounded-domain experiments use $[0,1]^D$.
Initial designs contain
$n_0=20,40,60,80,100$ Sobol points for $D=2,\ldots,6$.
Every sequential method selects the argmax over the same Sobol candidate
pool.
Observation noise is fixed at $5\%$ of the function standard deviation,
estimated once from a pilot sample.

The primary evaluator is the parameter-free $1$-NN reconstruction described
in Appendix \ref{app:evaluator}.
All methods use the same initial designs within each paired seed.

\paragraph{Acquisition methods.}
\begin{enumerate}[leftmargin=2em,itemsep=1pt,topsep=2pt]
    \item \emph{FPS}: candidate farthest from the current design.
    \item \emph{Sobol} and \emph{Random}: fixed designs independent of
          observed values.
    \item \emph{EIG}: posterior-variance maximization under the fixed
          acquisition kernel.
    \item \emph{EIG$\times g$}: EIG with the geometric equalizer.
    \item \emph{$g$ + reconstruction warp}: the full method.
          The warp-only arm drops $g$.
    \item \emph{$g$ + Kumaraswamy}: the trained input-warp baseline of
          Appendix \ref{app:warp_details}.
    \item \emph{NIPV} and \emph{EPIG}: integrated covariance-based criteria
          of Appendix \ref{app:zoo_criteria}.
    \item \emph{Refit}: variance acquisition with the acquisition-kernel
          hyperparameters re-estimated each round.
\end{enumerate}

\paragraph{Neumann acquisition kernel.}
For a one-dimensional stationary kernel $\kappa$, the reflected Neumann
construction of \citet{bankestad2026boundary} is approximated by
\begin{equation}
    k_N(x,x')
    =
    \sum_{n=-L}^{L}
    \left[
        \kappa(x-x'+2n)
        +
        \kappa(x+x'+2n)
    \right].
    \label{eq:neumann_1d}
\end{equation}
For product kernels on $[0,1]^D$, the construction is applied dimension-wise
and normalized to unit diagonal.

\paragraph{Lengthscale study.}
The lengthscale experiment of \Cref{fig:cube_ell} evaluates fixed
acquisition lengthscales spanning $0.5$--$2$ times the design-spacing value,
together with a per-round refitted variant.
All methods use the same initial designs, candidate pools, and observation
noise.

\section{All benchmark trajectories}
\label{app:all_curves}

\begin{figure}[ht]
    \centering
    \includegraphics[width=\linewidth]{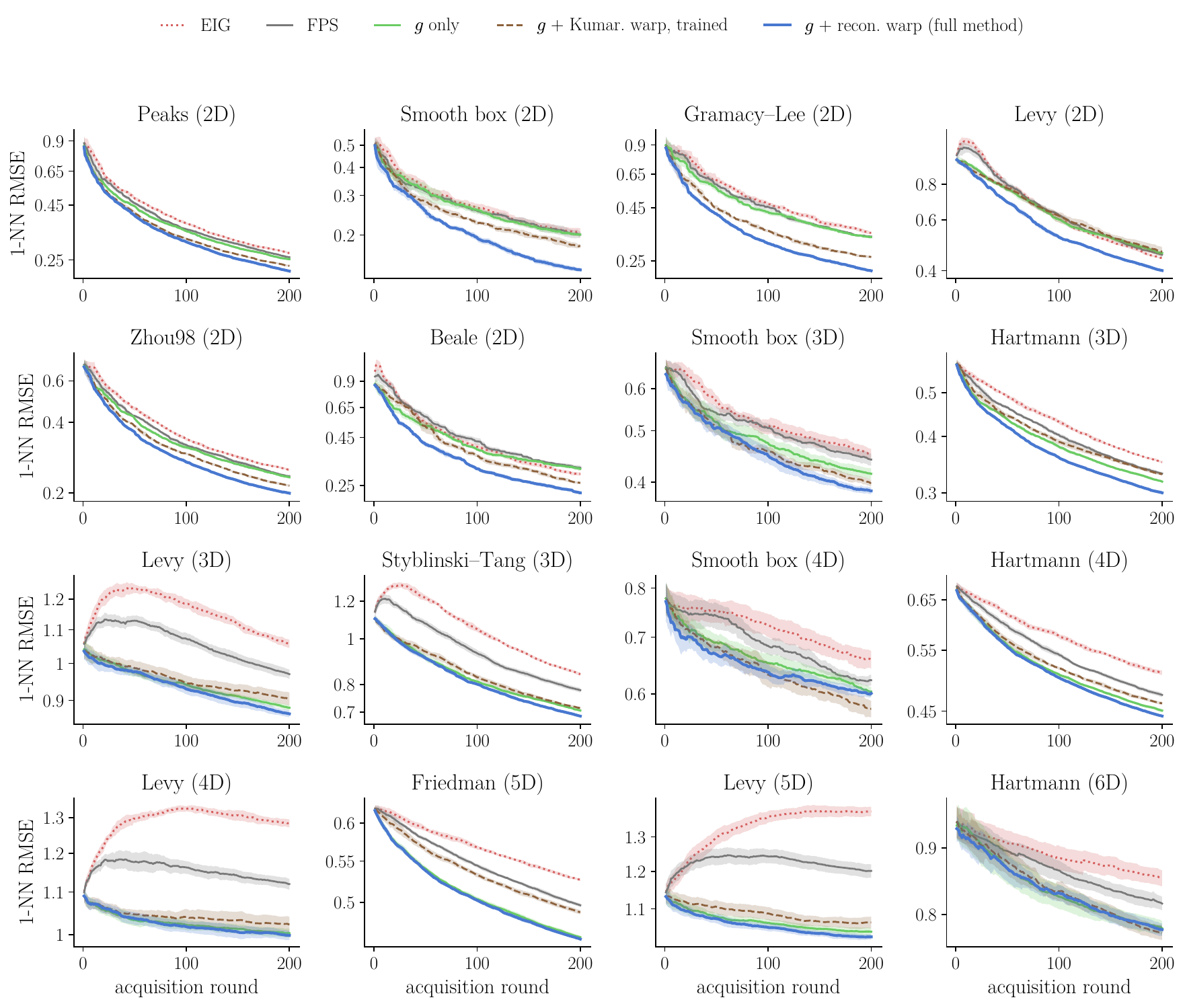}
    \caption{
    Reconstruction error over acquisition rounds for all sixteen synthetic
    benchmarks (mean and $95\%$ confidence interval over $10$ seeds).
    The eight benchmarks shown in \Cref{fig:main_curves} are a subset.}
    \label{fig:all_curves}
\end{figure}

\Cref{fig:all_curves} reports the complete reconstruction-error trajectories
for all sixteen synthetic benchmarks.

\section{Robustness to the reconstruction evaluator}
\label{app:evaluator}

The primary evaluator is intentionally independent of the model used for
acquisition.
Given design locations $\{\vx_i\}_{i=1}^{n}$ and target function $f$, we
measure the mean squared error of the nearest-neighbor reconstruction on a
dense held-out set $T$:
\begin{equation}
    \mathrm{MSE}_{1\text{-NN}}
    =
    \frac1{|T|}
    \sum_{\vx\in T}
    \left(
        f(\vx_{\mathrm{nn}(\vx)})
        -
        f(\vx)
    \right)^2,
    \qquad
    \mathrm{nn}(\vx)
    =
    \arg\min_i\|\vx-\vx_i\|.
    \label{eq:onenn}
\end{equation}

The evaluator has no kernel, fitted lengthscale, or uncertainty model, and
therefore applies equally to GP-based methods, FPS, and fixed designs.
It also matches the reconstruction loss underlying the design-density derivation in \Cref{sec:design_density}, making robustness to alternative evaluators an important check.

We therefore rescore every acquired design with $3$-NN, $5$-NN, and a
fixed-lengthscale Mat\'ern-$5/2$ GP reconstruction.

\begin{figure}[t]
    \centering
    \includegraphics[width=\linewidth]{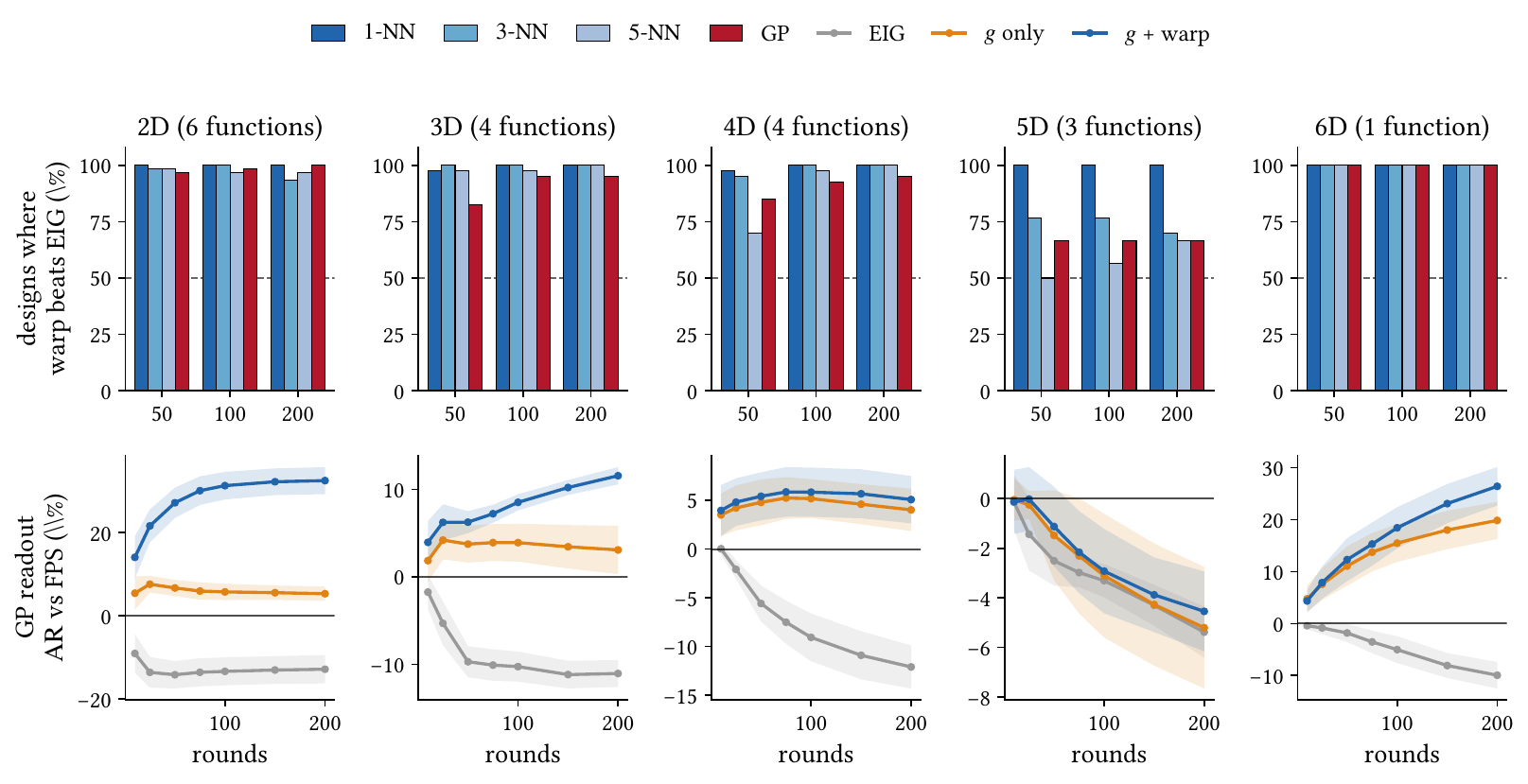}
    \caption{
    Robustness to the reconstruction evaluator across the sixteen synthetic
    and two real-data benchmarks.
    Top: fraction of designs for which the full method improves on unwarped
    EIG under each evaluator and at three acquisition horizons.
    Bottom: area reduction relative to FPS under the GP evaluator
    (mean and $95\%$ confidence interval over $10$ seeds).}
    \label{fig:readout_ar}
\end{figure}

\Cref{fig:readout_ar} shows that the main qualitative conclusions persist
across reconstruction evaluators.
At $100$ rounds, the warp improves on unwarped EIG for $100\%$ of designs
under $1$-NN, $96\%$ under $3$-NN, $91\%$ under $5$-NN, and $91\%$ under the
GP readout.
Under the GP evaluator, the full method improves on FPS on sixteen of the
eighteen benchmarks.

The main evaluator-dependent effect is the contribution of the geometric
equalizer.
A nearest-neighbor reconstruction does not extrapolate beyond the acquired
design, whereas a stationary GP reverts toward its prior in sparsely sampled
regions.
Moving design points away from the boundary therefore carries a larger penalty
under the GP readout for functions whose boundary behavior is difficult to
extrapolate.
This explains why the equalizer contributes less under the GP evaluator
than under $1$-NN, while the benefit of the reconstruction warp is more
stable across evaluators.

\section{Area reduction}
\label{app:area_reduction}

Let $M_t$ denote the reconstruction error after acquisition round $t$.
For an acquisition horizon $T$, define
\begin{equation}
    A
    =
    \sum_{t=1}^{T} M_t.
\end{equation}
Area reduction relative to the FPS trajectory from the same initial design is
\begin{equation}
    \mathrm{AR}
    =
    \frac{
        A^{(\mathrm{FPS})}
        -
        A^{(\mathrm{method})}
    }{
        A^{(\mathrm{FPS})}
    }.
    \label{eq:area_reduction}
\end{equation}
AR is reported in percent, with positive values indicating lower cumulative
reconstruction error than FPS.
It is computed per paired seed before averaging across seeds or benchmarks.

\section{Benchmark functions}
\label{app:test_functions}

All synthetic functions are evaluated on inputs scaled to $[0,1]^D$ and
mapped internally to their native domains.

\paragraph{Gramacy--Lee 2D~\citep{gramacy2012cases}.}
On $[-2,6]^2$,
\[
    f(x_1,x_2)
    =
    x_1\exp(-x_1^2-x_2^2).
\]

\paragraph{Peaks 2D.}
On $[-3,3]^2$,
\[
\begin{split}
    f(x,y)
    ={}&
    3(1-x)^2e^{-x^2-(y+1)^2}
    -
    10\left(\frac{x}{5}-x^3-y^5\right)e^{-x^2-y^2}\\
    &-
    \frac13 e^{-(x+1)^2-y^2}.
\end{split}
\]

\paragraph{Smooth box, $D=2,3,4$.}
\label{app:box}
On the native domain $[-3,3]^D$,
\begin{equation}
    f(\vx)
    =
    \prod_{i=1}^{D}
    \sigma\!\bigl(k(x_i-b_{\mathrm{start}})\bigr),
    \label{eq:smooth_box}
\end{equation}
with $k=20$ and $b_{\mathrm{start}}=-0.5$.

\paragraph{Hartmann, $D=3,4,6$.}
We use the standard Hartmann functions~\citep{surjanovic2013virtual},
\[
    f(\vx)
    =
    -
    \sum_{i=1}^{4}
    \alpha_i
    \exp
    \left(
        -\sum_{j=1}^{D}
        A_{ij}(x_j-P_{ij})^2
    \right),
\]
with the standard coefficient matrices and
$\alpha=(1.0,1.2,3.0,3.2)$.

\paragraph{Levy, $D=2,3,4,5$~\citep{surjanovic2013virtual}.}
On $[-10,10]^D$, let
$w_i=1+(x_i-1)/4$ and define
\[
\begin{split}
    f(\vx)
    ={}&
    \sin^2(\pi w_1)
    +
    \sum_{i=1}^{D-1}
    (w_i-1)^2
    \left[
        1+10\sin^2(\pi w_i+1)
    \right]\\
    &+
    (w_D-1)^2
    \left[
        1+\sin^2(2\pi w_D)
    \right].
\end{split}
\]

\paragraph{Zhou (1998) 2D~\citep{surjanovic2013virtual}.}
With $\phi$ denoting the standard multivariate normal density,
\[
    f(\vx)
    =
    \frac{10^2}{2}
    \left[
        \phi\!\left(10(\vx-\tfrac13)\right)
        +
        \phi\!\left(10(\vx-\tfrac23)\right)
    \right].
\]

\paragraph{Beale 2D~\citep{surjanovic2013virtual}.}
On $[-4.5,4.5]^2$,
\[
    f(x,y)
    =
    (1.5-x+xy)^2
    +
    (2.25-x+xy^2)^2
    +
    (2.625-x+xy^3)^2.
\]

\paragraph{Styblinski--Tang 3D~\citep{surjanovic2013virtual}.}
On $[-5,5]^3$,
\[
    f(\vx)
    =
    \frac12
    \sum_{i=1}^{3}
    \left(
        x_i^4-16x_i^2+5x_i
    \right).
\]

\paragraph{Friedman 5D~\citep{friedman1991mars}.}
On $[0,1]^5$,
\[
    f(\vx)
    =
    10\sin(\pi x_1x_2)
    +
    20(x_3-0.5)^2
    +
    10x_4
    +
    5x_5.
\]

\subsection{Real-data benchmarks}
\label{app:real_test_functions}

We use two UCI regression datasets:
Combined Cycle Power Plant
($9{,}568$ observations, four inputs)
and Airfoil Self-Noise
($1{,}503$ observations, five inputs).

To obtain a continuous reference target from each finite dataset, we fit a Mat\'ern-$5/2$ ARD GP to the full dataset after min--max scaling the inputs to $[0,1]^D$.
The posterior mean of this GP is used only as the evaluation target and is
never available to the acquisition methods.

The fitted surrogates achieve held-out
$R^2=0.95$ for the power-plant data and
$R^2=0.92$ for airfoil self-noise.
All acquisition methods are evaluated against the same fixed surrogate
target.

\end{document}